\documentclass[a4paper,fleqn]{cas-dc}

\usepackage[numbers]{natbib}
\usepackage{makecell}
\usepackage{amsmath}
\usepackage{color}
\usepackage{soul}
\usepackage{subfigure}

\def\tsc#1{\csdef{#1}{\textsc{\lowercase{#1}}\xspace}}
\tsc{WGM}
\tsc{QE}
\tsc{EP}
\tsc{PMS}
\tsc{BEC}
\tsc{DE}

\begin{document}
\begin{sloppypar}
\let\WriteBookmarks\relax
\def\floatpagepagefraction{1}
\def\textpagefraction{.001}
\shorttitle{BIMHA for Video Sentiment Analysis}
\shortauthors{Ting Wu et~al.}

\title [mode = title]{Video Sentiment Analysis with Bimodal Information-augmented Multi-Head Attention}

\author[1]{Ting Wu}[type=editor,
                        auid=000,bioid=1,
                        orcid=0000-0002-0538-0255]

\ead{wuting1972@shu.edu.cn}
\credit{Conceptualization of this study, Methodology, Software, Writing - Original Draft}
\address[1]{School of Computer Engineering and Science, Shanghai University, Shanghai, China}

\author[1,2]{Junjie Peng}[orcid=0000-0001-9091-8306]
\cormark[2]
\ead{jjie.peng@shu.edu.cn}
\credit{Conceptualization of this study, Writing - Review \& Editing, Supervision}
\address[2]{Shanghai Institute for Advanced Communication and Data Science, Shanghai University, Shanghai, China}

\author[3,4]{Wenqiang Zhang}[ ]
\cormark[1]
\ead{wqzhang@fudan.edu.cn}
\credit{Conceptualization of this study, Writing -  Review \& Editing}
\address[3]{Academy for Engineering \& Technology, Fudan University, Shanghai, China}
\address[4]{School of Computer Science and Technology, Fudan University, Shanghai, China}

\author[1]{Huiran Zhang}[]
\credit{Conceptualization of this study, Resources}

\author[5,6]{Shuhua Tan}[]
\credit{Conceptualization of this study, Resources}
\address[5]{National Engineering Laboratory for Logistics Information Technology, Shanghai, China}
\address[6]{YTO Express Co., Ltd., Shanghai, China}

\author[5,6]{Fen Yi}[]
\credit{Conceptualization of this study, Resources}

\author[1]{Chuanshuai Ma}[]
\credit{Validation, Investigation}

\author[1]{Yansong Huang}[]
\credit{Formal analysis, Visualization}



\begin{abstract}
Humans express feelings or emotions via different channels. Take language as an example, it entails different sentiments under different visual-acoustic contexts. To precisely understand human intentions as well as reduce the misunderstandings caused by ambiguity and sarcasm, we should consider multimodal signals including textual, visual and acoustic signals. The crucial challenge is to fuse different modalities of features for sentiment analysis. To effectively fuse the information carried by different modalities and better predict the sentiments, we design a novel multi-head attention based fusion network, which is inspired by the observations that the interactions between any two pair-wise modalities are different and they do not equally contribute to the final sentiment prediction. By assigning the acoustic-visual, acoustic-textual and visual-textual features with reasonable attention and exploiting a residual structure, we attend to attain the significant features. We conduct extensive experiments on four public multimodal datasets including one in Chinese and three in English. The results show that our approach outperforms the existing methods and can explain the contributions of bimodal interaction in multiple modalities.
\end{abstract}




\maketitle

\section{Introduction}
The rapid development of human society not only promotes the upgrading of science and technology, but also puts forward higher requirements for the intelligence of current service industries. Since working hours, labor demand and service quality are three important issues that need to solve in this area, many automated robots such as intelligent customer service, intelligent shopping guide and intelligent escort robot are designed to help people do some works. Although the first two problems can be solved to a large extent by using these robots, the problem of lacking understanding of human intentions still exists, which results in poor service quality and low user satisfaction \cite{Huetal2020,Huetal2021}. It is therefore essential to understand human intentions for making reasonable replies and behaviors.

Sentiment analysis is one of the most crucial techniques for correctly understanding people's intentions. From the perspective of cognition, human learns words not just by semantic or syntactic accompaniments, but also by visual and acoustic reinforcements \cite{Zadehetal2020}. That means human learning from real-world experience is usually multisensory. Judging sentiment can not rely solely on text, especially in today's society where people express opinions with videos.

As video-based sentiment analysis involves multimodal data, exploring connection and mining complementary information are huge challenges. To address such challenges, information fusion is important. It can aid to mimic the way humans process and analyze text and, hence, overcome the limitations of standard approaches to achieve computing and sentiment analysis \cite{Hussainetal2021}. In fact, different modalities reflect sentiment with various intensities, and they may possess consistent or independent information. To efficiently mine the sentiment tendency expressed by different modalities, attention mechanism has been widely used for multimodal fusion. For example, Mai et al. \cite{Maietal2020_a} proposed attention-guided fusion  focusing on the key time steps determined by acoustic information. Rahman et al. \cite{Rahmanetal2020} utilized attention mechanism to learn the nonverbal behaviors. Tsai et al. \cite{Tsaietal2019_b} designed the interactive module with attention to capture the cross-modal information.

Though attention mechanism has been widely studied, most of the existing methods utilize it to extract informative features of a single modality, and they fail to consider interactive information generated by inter-modal interaction differences. Generally, one modality can provide additional information for the other modality, and the fusion features of the two modalities contribute differently to the final sentiment decision. For instance, it can be inferred that a person is happy when he speaks loudly with a pleasant smile, but he may be angry when he speaks loudly with spoken content expressing dissatisfaction. Tzirakis et al.\cite{Tzirakisetal2021} visualized the hierarchical attention and found that the proposed model assigns the audiovisual stream the highest scores in every layer. It indicates that exploring the interaction between different modalities is meaningful and well-founded. And it is necessary to find an efficient method to weigh the information provided by the interaction of pair-wise modalities so that computers can accurately recognize human sentiment.
 
In fact, some hierarchical fusion based methods have considered the bimodal interaction \cite{Hanetal2021,Huddaretal2020_a,Maietal2020_b,Majumderetal2018}. In hierarchical fusion, bimodal interactions are generated based on unimodal information, and trimodal interactions are obtained based on bimodal information. The method proposed by Majumder et al. uses fully connected layers followed by GRU to perform fusion between different modalities and incorporate contextual information. In each fusion stage, the input is the concatenation of unimodal, bimodal features, respectively \cite{Majumderetal2018}. The attention-based inter-modal fusion proposed by Huddar et al. utilizes soft attention and element-wise matrix multiplication to get the bimodal attention matrices, which are then concatenated as the trimodal attention matrix for the final classification. In this method, bidirectional relationships in each modality pair are modeled as different forms. For instance, the representations of acoustic-visual and visual-acoustic are different \cite{Huddaretal2020_a}. The Bi-Bimodal Fusion Network proposed by Han et al. devotes to balancing the contribution of different modality pairs properly. But it just takes two text-related modality pairs, textual-visual and textual-acoustic as the inputs \cite{Hanetal2021}. The Adversarial Representation Graph Fusion framework (ARGF) proposed by Mai et al. focuses on joint embedding space learning and adopts a novel Graph Network to explore unimodal, bimodal, and trimodal dynamics successively \cite{Maietal2020_b}. These studies break through the conventional ways to model hierarchically. However, there are some challenges for introducing bimodal interaction to promote multimodal fusion. As the interactions between every two pair-wise modalities are different, simple concatenation fails to mine the inter-bimodal relationships. For each sample, the importance of acoustic-visual, acoustic-textual, and visual-textual may vary a lot.
 
In this paper, we study the contribution of pair-wise modalities for video sentiment analysis. Meanwhile, we devote to considering intra-modal as well as inter-modal and inter-bimodal interaction. Further, we develop a bimodal information oriented multi-head attention based architecture to extract the independent and consistent information from different modalities for effective fusion. Specifically, the model obtains interaction information between modalities by tensor fusion and enhances the bimodal interaction with the extended multi-head attention mechanism. By calculating bimodal attention in different feature subspaces, the obtained weighted bimodal features are fused with the original inter-modal features. And the fused results are used as the input of the sentiment inference network. We evaluate the effectiveness and generalization of the proposed approach with extensive experiments on CH-SIMS \cite{Yuetal2020} dataset, CMU-MOSI \cite{Zadehetal2016} dataset, MOSEI \cite{Zadehetal2018_b} dataset, and IEMOCAP \cite{Bussoetal2008} dataset. The results show that our model outperforms the existing methods. Besides, we intuitively explain the principle of the multi-head attention method based on bimodal information-augmented, and discuss the contribution of each two pair-wise modalities among multiple modalities.

The main contributions can be summarized as follows:
\begin{itemize} \item We propose a multi-head attention based model for video based sentiment analysis which follows hierarchical fusion and considers interactions in intra-modal, inter-modal and inter-bimodal. We call the extended multi-head attention Bimodal Multi-Head Attention (BMHA), which contains three traditional Multi-Head Attention (MHA). Each of them utilizes multimodal features as the source information and bimodal features as the target information. \item We explore the relative importance and relationships between any two pair-wise modalities and provide visualized explanations on the differences of the interactions. Besides, we find the performance of BMHA can be enhanced to some extent with bimodal information. Bimodal attention is proved to be efficient for performing sentiment prediction. \item We conduct extensive experiments on four public datasets, one is in Chinese and the others are in English, and provide new benchmark results in multimodal sentiment analysis.
\end{itemize}

\section{Related Work}

Video is a kind of temporal data mixed with multiple modalities. Processing multimodal data needs to perform multimodal embeddings learning, which turns to model intra-modal and cross-modal dynamics \cite{Gkoumasetal2021}. Early, late, and hybrid fusion are the common strategies used for modeling such dynamics. Early fusion approaches integrate features after being extracted \cite{Williamsetal2018}. Late fusion approaches build models for each modality and then combine their decisions by averaging, weighted sum, majority voting, or deep neural networks \cite{Sarahetal2021}. Hybrid strategies combine outputs from early fusion and individual unimodal predictions \cite{Lietal2020}.
 
As a powerful machine learning technique, deep learning is widely used to construct sophisticated approaches and achieves satisfying results in multimodal analysis. In the early works, Zadeh et al. \cite{Zadehetal2017} proposed the tensor-based network, Tensor Fusion Network (TFN), based on the tensor product of modalities. Liu et al. \cite{Liuetal2018} proposed Low-rank Multimodal Fusion (LMF) to reduce the high dimensionality of TFN and improve the efficiency of the model. As human language contains time-varying signals, researchers proposed models based on CNN, RNN and LSTM to capture the interactions over time steps \cite{HochreiterSchmidhuber1997,Huddaretal2020_b,Maietal2020_c,Neumannvu2017}. Gkoumas et al. \cite{Gkoumasetal2021}  decomposed the fusion problem into multiple stages and conducted Recurrent Multistage Fusion Network (RMFN) with different number of stages needed to model the cross-modal dynamics. Mai et al. \cite{ Maietal2019} put forward Hierarchical Feature Fusion Network (HFFN) to model the local and global interactions through a divide-and-conquer approach. Some researchers introduced the memory unit to model modality interactions which is the extension of recurrent neural model. Zadeh et al. \cite{Zadehetal2018_a,Zadehetal2018_b} proposed the Memory Fusion Network (MFN), Dynamic Fusion Graph (DFG). MFN builds a multimodal gated memory component whose memory cell is updated along with the evolution of the hidden states of three unimodal LSTMs. DFG replaces the module in MFN with a new neural-based component.

Inspired by the approaches for the machine translation task in Natural Language Processing (NLP), encoder-decoder structures in sequences to sequences learning are introduced. They model human language by converting one source modality to another target modality. Pham et al. \cite{Phametal2019} proposed the Multimodal Cyclic Translations Network (MCTN) which is a hierarchical neural machine translation network with one source modality and two target modalities for learning joint embedding. Tsai et al. \cite{Tsaietal2019_b} transformed source modality to the target modality using directional pairwise cross-modal transformers. To reduce the modality gap,  Mai et al.  \cite{Maietal2020_b} constructed the ARGF to translate the distributions of source modalities into that of the target modalities using adversarial training. Wang et al. \cite{Wangetal2020} exploited a parallel translation approach that fuses linguistic with acoustic features and linguistic with visual features independently to eliminate noisy interference between modalities. Some holistic frameworks endow models with inherent interpretability by separating cross-modal interactions. For example, Zadeh et al. \cite{Zadehetal2019} applied seven distinct self-attention mechanisms to the multimodal representation, capturing all possible unimodal, bimodal, and trimodal interactions, simultaneously. Hung et al. \cite{Hungetal2020} used routing algorithm to dynamically model the contribution of unimodal, bimodal and trimodal explanatory features for predictions in a local manner. Hazarika et al. \cite{Hazarikaetal2020} introduced two subspaces, a joint subspace and a modality-specific subspace, to capture unimodal and trimodal interactions.

Since Seq2Seq models fail to accommodate long sentences and the fixed-size context vectors can not encapsulate information from longer sentences, attention mechanism is introduced in the translation task. It can weigh the influences and extract important features \cite{YuJiangetal2020}. Besides feature extraction, it is popular in multimodal fusion. Huddar et al. \cite{Huddaretal2020_a} calculated the bimodal attention matrix representations separately, and concatenated them as the trimodal attention matrix to fuse the interaction information from different modalities. Xu et al. \cite{Xuetal2020} proposed head fusion, a multi-head self-attention method. Tzirakis et al. \cite{Tzirakisetal2021} proposed an end-to-end multimodal model for affective recognition, which performs hierarchical attention to fuse the modality-specific features. Kim et al. replaced the attention module in \cite{Poriaetal2017} with the Scaled Dot-Product Attention to calculate the attention score of each modality and used the multi-head attention mechanism to learn features in multiple representation subspaces at different positions \cite{Kimlee2020,Vaswanietal2017}. Xi et al. \cite{Xietal2020} proposed a method based on the multi-head attention mechanism, which uses the self-attention mechanism to extract the intra-modal features and employs the multi-head mutual attention to analyze the correlation between different modalities.

Other strategies involve interesting domains with novel ideas including reinforcement learning \cite{BroekensChetouani2021,Chenetal2017, Pengetal2021, Zhangetal2021}, fuzzy logic \cite{Chaturvedietal2019}, bilinear pooling \cite{Zhangetal2019}, deep canonical correlation analysis \cite{Sunetal2020}, domain adaptation \cite{Qietal2018}, topic recognition \cite{Stappenetal2021}, modality distillation \cite{PengHZC2021}, quantum learning\cite{Lietal2021,Zhangetal2020,Zhangetal2018},  and special physiological signals like body postures, EEG \cite{Moghimietal2020,Noroozietal2021}.

\section{Method}

In order to solve the problem that different interaction between multiple modalities makes the information contribution different, a video sentiment analysis model using Bimodal Information-augmented Multi-Head Attention (BIMHA) is put forward. In this model, the text (T), audio (A) and video (V) are taken as the input. Besides, the main two parts are inter-modal interaction and inter-bimodal interaction. For the inter-modal interaction, previous works of \cite{Liuetal2018,Zadehetal2017} on multi-modal feature fusion have shown that the outer product can learn interactions between different features effectively. Thus, we use the outer product to represent visual-textual (VT), acoustic-textual (AT) and acoustic-visual (AV) features. For the inter-bimodal interaction, an extended multi-head attention mechanism is designed to calculate bimodal attention. Merging the inter-modal features and inter-bimodal features, we conduct the final sentiment prediction.

\begin{figure*}
	\centering
	\includegraphics[scale=.35]{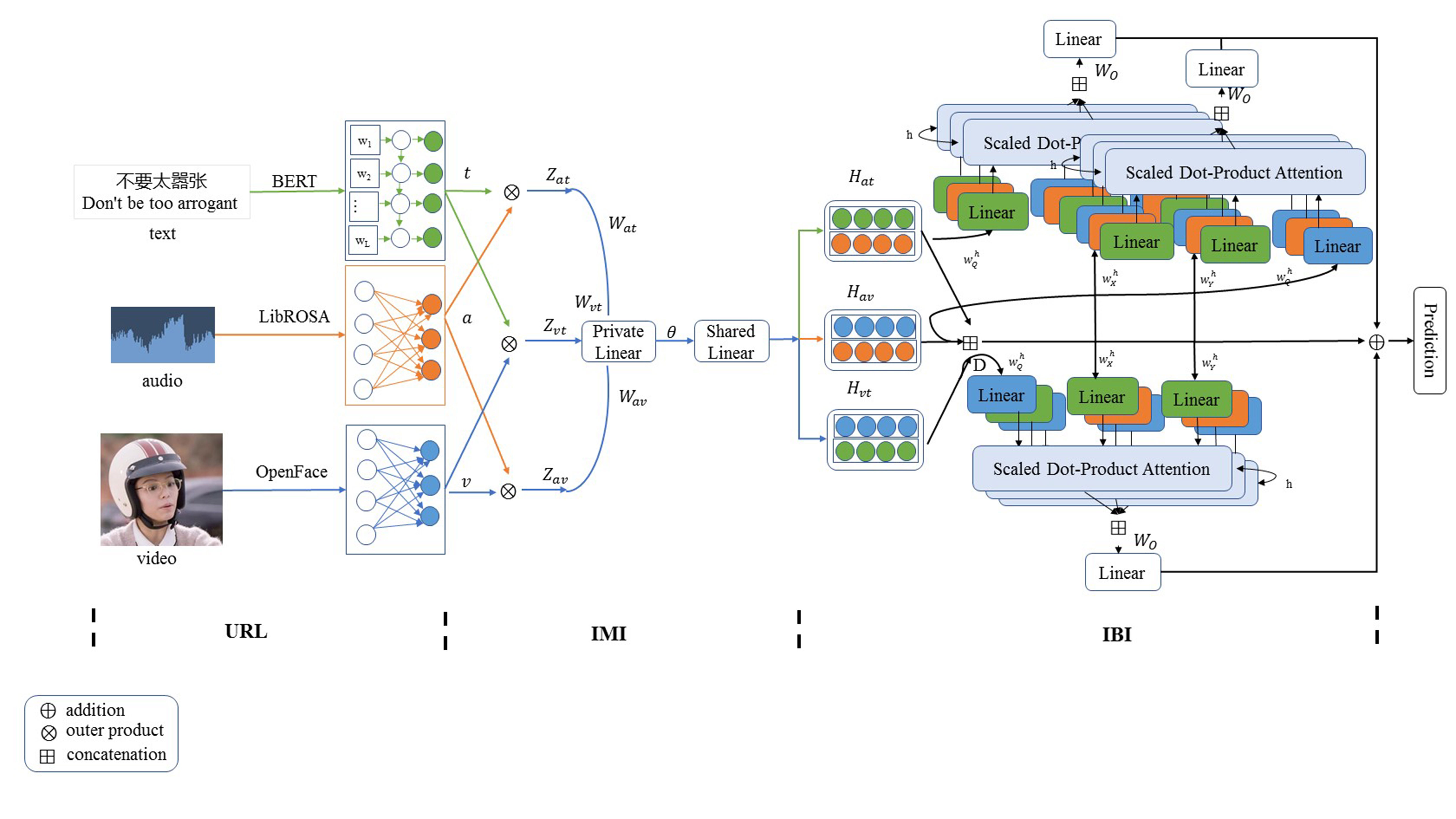}
	\caption{The network architecture of BIMHA. It consists of four components: URL for extracting unimodal features; IMI for modeling inter-modal interaction; IBI for learning inter-bimodal interaction; The last for sentiment prediction. The input of the model includes text, audio and image sequences.}
	\label{FIG:1}
\end{figure*}

The architecture of the model is shown in Figure 1. From Figure 1, it is easy to notice that the model consists of four parts, i.e., Unimodal Representation Learning (URL), Inter-Modal Interaction (IMI), Inter-Bimodal Interaction (IBI) and prediction network. Among these parts, URL is to model modality-specific interactions. IMI is to obtain the interaction information between every two modalities. IBI is to learn bimodal interactions between every two pair-wise modalities, where Bimodal Multi-Head Attention (BMHA) is designed to obtain the key features assigned with large weights. Since the output features from IMI have different dimensions, these features are input into two linear layers to adapt to the IBI module. The first fully connected layer is private for converting to the unified dimensions $d$, while the second is to extract deep features and reduce parameters via the sharing mechanism. Finally, the obtained interaction information and the original information serve as the input of the prediction layer to get the final sentiment label.

\subsection{Unimodal Representation Learning}

Videos in the dataset are split into small utterances. Each utterance contains three kinds of unimodal features, i.e., textual features, acoustic features and visual features. For textual modality, pre-trained Chinese BERT \cite{Devlinetal2019} is employed to get the $d_t$ -dimensional sentence embedding. In each sentence, the length of the word sequence is different. Thus we adopt padding and truncation to let the final length be $L$. $L$ is calculated by two steps. First, we obtain the average length of the sentences and calculate the standard deviation of raw lengths. Then we take the sum of the average and $\lambda$ times the standard deviation as the final length. Padding is introduced to fill the short sentence with the specific characters at the end. For the long sentence, the first $L$ vectors are taken to constitute the sentence embedding. To fully mine the semantic relationships between words in the sentences, the LSTM network is used to generate fusion features for each time step. And we employ the final hidden state output as the sentence embedding $t_i\in \mathbb R^{d_{t1}}$ with $d_{t1}$ dimensions.

For acoustic and visual modalities, LibROSA \cite{Mcfeeetal2015} is used to extract  $d_a$-dimensional acoustic features including Zero Crossing Rate (ZCR), Mel-Frequency Cepstral Coefficients (MFCCs) and Constant-Q chromatogram (CQT). And Ffmpeg\footnote{https://ffmpeg.org/ffmpeg.html} is used to frame the videos at a certain rate and MTCNN \cite{Zhangetal2016} is applied to extract the aligned faces. Based on these aligned faces, $d_{v}$-dimensional facial features are extracted by Multi Comp OpenFace2.0\footnote{https://github.com/TadasBaltrusaitis/OpenFace/wiki}. Then, acoustic features $a_i\in \mathbb R^{d_{a1}}$ with $d_{a1}$ dimensions and visual features $v_i\in \mathbb R^{d_{v1}}$ with $d_{v1}$ dimensions are further extracted with three-layer deep neural networks. Note that the number of frames in each audio or video clip is not always equal, so in the data preprocessing, we average all the features of the frames in each utterance.

\subsection{Inter-Modal Interaction Information}

With the unimodal features of each utterance obtained from section 3.1, the unimodal features of the whole dataset can be represented as $Z_t=\{t_1,t_2,...,t_N\}$,  $Z_a=\{a_1,a_2,...,\\a_N\}$,  $Z_v=\{v_1,v_2,...,v_N\}$, where $Z_t$, $Z_a$ and $Z_v$ denotes the set of textual features, acoustic features and visual features respectively, \emph{N} is the number of samples in the dataset. The tensor fusion of any two modalities is the outer product. As shown in equation (1), the AV feature matrix, AT feature matrix and  VT feature matrix can be learned based on $Z_t$, $Z_a$ and $Z_v$, where $d_{av} = d_{a1} \times d_{v1}$,  $d_{at} = d_{a1} \times d_{t1}$ and  $d_{vt} = d_{v1} \times d_{t1}$.
\begin{flalign}
&Z_{av}=Z_a\otimes Z_v, Z_{av} \in \mathbb R^{d_{av}}& \nonumber\\
&Z_{at}=Z_a\otimes Z_t, Z_{at} \in \mathbb R^{d_{at}}& \\
&Z_{vt}=Z_v\otimes Z_t, Z_{vt} \in \mathbb R^{d_{vt}}& \nonumber
\end{flalign}
In order to adapt to the calculation of the IBI module, two fully connected layers with $d$ units are used to transform these features. For the private linear layers, they act on AV, AT and VT features with ReLU activation function. The transformed features are shown in equation (2), where $W_{av} \in \mathbb R^{d_{av} \times d}$, $W_{at} \in \mathbb R^{d_{at} \times d}$ and $W_{vt} \in \mathbb R^{d_{vt} \times d}$ are learnable transformation matrices, $b_{av}$, $b_{at}$ and $b_{vt}$ are biases.
\begin{flalign}
&\bar{Z}_{av}= ReLU(W_{av} \times Z_{av}+b_{av})& \nonumber\\
&\bar{Z}_{at}= ReLU(W_{at} \times Z_{at}+b_{at})& \\
&\bar{Z}_{vt}= ReLU(W_{vt} \times Z_{vt}+b_{vt})& \nonumber
\end{flalign}

After the transformation of the private layer, the features with consistent dimensions are obtained. In order to further extract the deep features, $\bar{Z}_{av}$, $\bar{Z}_{at}$,  and $\bar{Z}_{vt}$ are input into the shared layer. The shared layer means that the parameters for training these three pair-wise features are shared to reduce the storage space. As shown in equation (3), the inter-modal interaction features are obtained which are represented as $H_{av}\in \mathbb{R}^d$, $H_{at}\in \mathbb{R}^d$, $H_{vt}\in \mathbb{R}^d$, where  $s \in \{av, at, vt\} $, $FC$ is fully connected layer and $\theta$ denotes the learnable parameter matrix.
\begin{flalign}
& H_{s}=FC( \bar{Z}_{s},\theta) \in \mathbb R^{d}&
\end{flalign}

\subsection{Inter-Bimodal Interaction Information}

The essence of the attention function can be described as the mapping of the query (\emph{Q}) to a series of key (\emph{X})-value (\emph{Y}) pairs. In NLP, the common setting of the attention mechanism is that the key and value are the same. Based on this, MHA introduces scaled dot product and multi-head calculation which can capture relevant information in different feature subspaces. Because of the advantages, it is widely applied to many other NLP tasks. More specifically, in MHA, the \emph{Q}, \emph{X} and \emph{Y} of each head are first processed by the linear transformation layers, as shown in equation (4) where $W_Q$, $W_X$, $W_Y$ are the parameter matrices of \emph{Q}, \emph{X} and \emph{Y}, \emph{h} means the specific head. The scaled dot product attention is calculated as shown in equation (5), where $d$ is the dimension of \emph{X}. The attention scores of all heads are concatenated as the input of a linear transformation to obtain the value of multi-head attention, as shown in equation (6), where $W_O$ is the parameter matrix, $n$ is the number of heads.
\begin{flalign}
&\bar{X}^{h}=X \times W_X^h& \nonumber\\
&\bar{Q}^{h}=Q \times W_Q^h &\\
&\bar{Y}^{h}=Y \times W_Y^h & \nonumber
\end{flalign}
\begin{flalign}
&A^{h}=softmax(\frac{\bar{Q}^{h}\times(\bar{X}^{h})^T}{\sqrt{d}})\times \bar{Y}^{h}&
\end{flalign}
\begin{flalign}
&MHA(Q,X,Y)= [A^1; A^2; ... ;A^n]\times W_O&
\end{flalign}

In order to carry out bimodal interaction, calculate bimodal contribution and capture relevant information from different representation subspaces, the inter-modal features extracted in section 3.2 are concatenated as demonstrated in equation (7) . $D$ represents the multimodal features including all the features of multiple modalities. Based on this, BMHA aims to learn bimodal attention. As shown in Figure 2, the column on the right shows the scores for different colors, and in this case, $H_{at}$ is more important, followed by $H_{vt}$, then $H_{av}$. Given that pair-wise features have different significance with respect to the multimodal features, the BMHA's mission is to focus on these differences and figure out their proportionality. On the basis of MHA, BMHA contains three MHA with the slightly different form of inputs. Specifically, multimodal features are set as the source, while the features of AV, AT and VT are set as the target respectively. In other words, $D$ acts as the key and value, while the single bimodal feature acts as the query.
\begin{flalign}
&D=Concat(H_{av},  H_{at}, H_{vt})\in \mathbb{R}^{3d}&
\end{flalign}
\begin{figure}
	\centering
	\includegraphics[scale=.35]{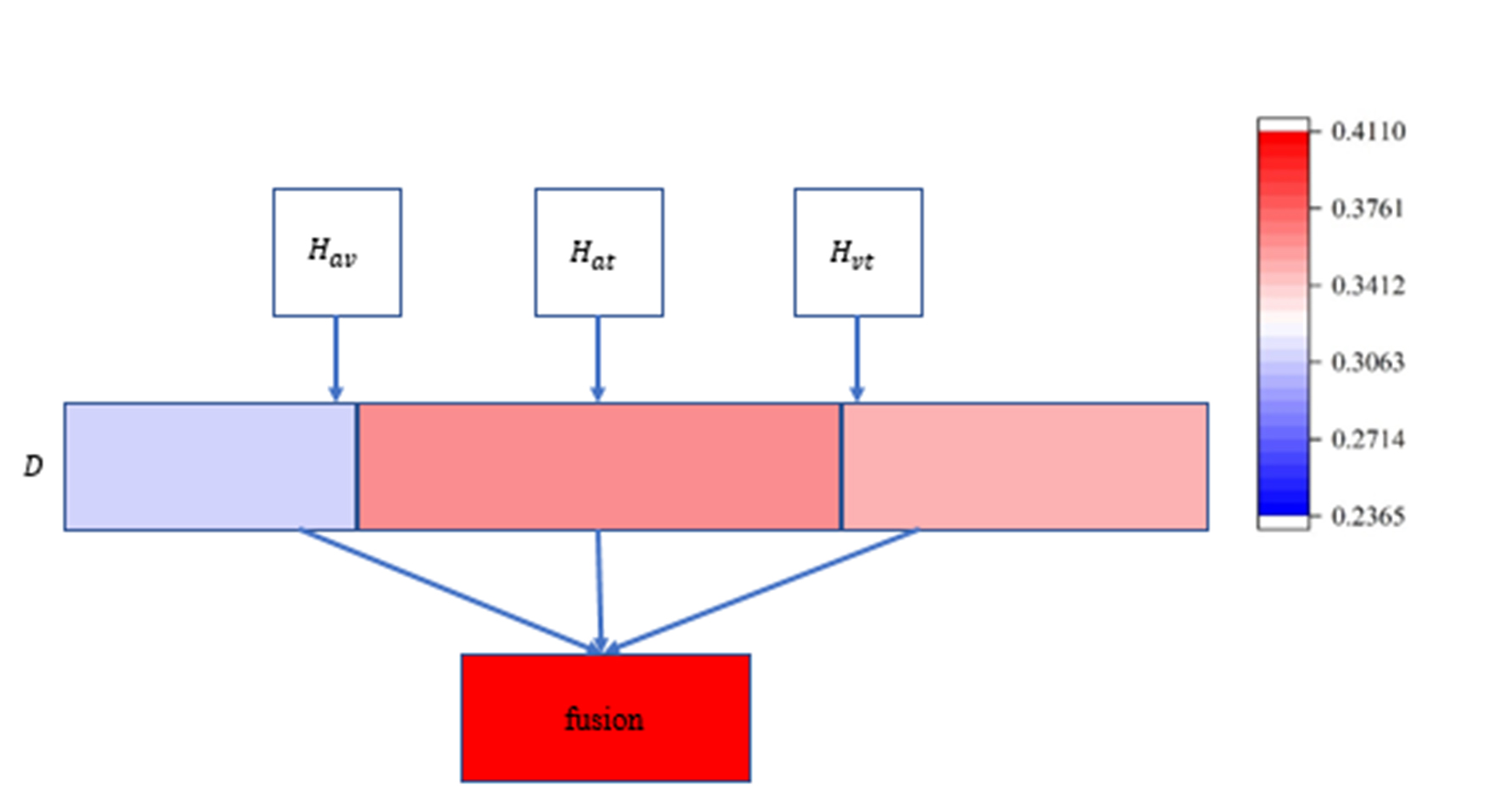}
	\caption{BMHA aims to capture pair-wise features that are more important in the fusion process. Our model focuses on assigning bimodal attention.}
	\label{FIG:2}
\end{figure}
First, we apply the multi-head linear projection on the feature matrices, i.e., $D$, $H_{av}$, $H_{at}$ and $H_{vt}$, and map them to the spaces (with the same number of dimensions $d_m$), as the following formulas show.
\begin{flalign}
&H_{D1}^i=W_{D1}^i\times D& \nonumber\\
&H_{D2}^i=W_{D2}^i\times D&\\
&\bar{H}_{s}^i=W_Q^i\times H_s, s \in \{av, at, vt\}& \nonumber
\end{flalign}
where $W_{D1} \in \mathbb R^{3d \times d_m}$, $W_{D2} \in \mathbb R^{3d \times d_m}$, $W_{Q} \in \mathbb R^{d \times d_m}$ are projection matrices for AV, AT, VT, $i$ is the index for operations in different projection space. Let the heads of BMHA be $h$, then $i \in \mathbb R^{3h}$. Note that we use the same parameter matrix ($W_Q^i$) for different bimodal features to reduce the number of parameters and the memory consumption. After obtaining the features in different projection spaces, we utilize the attention mechanism to explore the complementary relationships among the pair-wise modalities. The AV, AT, VT attention are applied as follows:
\begin{flalign}
&A_{av}^i=softmax(\frac{\bar{H}_{av}^i \times (H_{D1}^i)^T}{\sqrt{d_m}}) \times H_{D2}^i& \nonumber\\
&A_{at}^i=softmax(\frac{\bar{H}_{at}^i \times (H_{D1}^i)^T}{\sqrt{d_m}}) \times H_{D2}^i&\\
&A_{vt}^i=softmax(\frac{\bar{H}_{vt}^i \times (H_{D1}^i)^T}{\sqrt{d_m}}) \times H_{D2}^i& \nonumber
\end{flalign}

In order to obtain the bimodal feature representation with the attention assigned, the AV, AT and VT attentions of each head are respectively concatenated and subjected to the linear layer transformation. As shown in equation (10), $W_O \in \mathbb R^{hd\times d_m}$ is the weight parameter.
\begin{flalign}
&BMHA(H_{av}, D, D)= [A_{av}^1; ... ; A_{av}^h] \times W_O&\nonumber \\
&BMHA(H_{at}, D, D)= [A_{at}^1; ... ; A_{at}^h] \times W_O&  \\
&BMHA(H_{vt}, D, D)= [A_{vt}^1; ... ; A_{vt}^h] \times W_O& \nonumber
\end{flalign}

\subsection{Prediction}

The obtained inter-bimodal interaction features with $A_{av}$, $A_{at}$, $A_{vt}$ are concatenated and added as a residual function of the original features $D$, to avoid the vanishing gradient problem. Last, they are input into a three-layer DNN to generate the output.

\section{Experimental Settings}
\subsection{Datasets}

We use four public multimodal sentiment analysis datasets, CH-SIMS \cite{Yuetal2020}, CMU-MOSI \cite{Zadehetal2016}, MOSEI\cite{Zadehetal2018_b}, IEMOCAP \cite{Bussoetal2008}. The first one is in Chinese, and the others are in English. The statistics are shown in Table 1, and a brief introduction of them is as follows.

\noindent\textbf{CH-SIMS.} The CH-SIMS dataset contains 60 raw videos and 2281 refined video segments. The length of the clips is no less than one second and no more than ten seconds. In this dataset, the training, validation and test set are in the proportion 6:2:2. That is, there are 1368 utterances in the training set, 456 utterances in the validation set and 457 utterances in the test set. Besides, each utterance is annotated with the sentiment in the range [-1, 1], which corresponds to one of five categories, i.e., negative \{-1.0, -0.8\}, weakly negative \{-0.6, -0.4, -0.2\}, neutral \{0.0\}, weakly positive \{0.2, 0.4, 0.6\} and positive \{0.8, 1.0\}.

\noindent\textbf{CMU-MOSI.} The CMU-MOSI dataset contains 93 opinion videos collected from YouTube movie reviews. Each video is split into short segments and the final dataset consists of 2199 short monologue video clips with manual sentiment annotation in the range [-3, 3], which corresponds to the sentiment values ranging from highly negative to highly positive. Besides, there are 1284 utterances in the training set, 229 utterances in the validation set and 686 utterances in the test set.

\noindent\textbf{MOSEI.} The MOSEI dataset is a large dataset that contains 23453 annotated video segments from 1000 distinct speakers with 250 topics. Like MOSI, each video segment is manually annotated with sentiment value in the range [-3, 3], and the value refers to one sentiment category from strongly negative to strongly positive.

\noindent\textbf{IEMOCAP.} The IEMOCAP dataset contains dialogue videos from 10 actors. Each video is divided into segments with fine-grained emotional annotations including happy (H), sad (S), angry (A), neutral (N), excited, fearful, surprised, etc. In order to compare with other works, we only adopt the first four emotions as evaluation and calculate the Acc-2 and F1 score of each category in the dataset provided by \cite{Liuetal2018}.
\begin{table}[width=.9\linewidth,cols=4,pos=h]
	\caption{Datasets statistics in CH-SIMS, CMU-MOSI, MOSEI, and IEMOCAP}\label{tbl1}
	\begin{tabular*}{\tblwidth}{@{} LLLLLLLL@{} }
		\toprule
		Dataset & Train &  Valid & Test & Total \\
		\midrule
		CH-SIMS & 1368 & 456 & 457 & 2281\\
		CMU-MOSI & 1284 & 229 & 686 & 2199\\
		MOSEI & 16326 & 1871 & 4659 & 22856 \\
		IEMOCAP &2717 & 798 & 938 & 4453 \\
		\bottomrule
	\end{tabular*}
\end{table}

\subsection{Baselines}
\noindent\textbf{EF-LSTM.} The EF-LSTM \cite{Williamsetal2018} concatenates the original features of the three modalities and inputs them into the LSTM to capture the long-term dependencies between the modal sequences.

\noindent\textbf{LF-DNN.} The LF-DNN \cite{Yuetal2020} uses the DNN to learn unimodal features and then concatenates them as the input of the prediction layer.

\noindent\textbf{TFN.} The TFN \cite{Zadehetal2017} captures multimodal interaction information by creating a multi-dimensional tensor.

\noindent\textbf{LMF.} The LMF \cite{Liuetal2018} is the improvement of TFN, where low-rank multimodal tensors fusion technique is performed to improve the efficiency.

\noindent\textbf{MFN.} The MFN \cite{Zadehetal2018_a} stores the internal information of the modalities and the interaction information between the modalities through the gated memory unit and adds dynamic fusion graphs to reflect effective emotional information.

\noindent\textbf{DFG.} The DFG \cite{Zadehetal2018_b} replaces the fusion block in MFN with a Dynamic Fusion Graph, which is directly related to how modalities interact.

\noindent\textbf{MulT.} The MulT \cite{Tsaietal2019_b} uses its cross-modal attention module to extract the key information inside each modality and then merges these features based on the Transformer \cite{Vaswanietal2017} model.

\noindent\textbf{MISA.} The MISA \cite{Hazarikaetal2020} incorporates the combination of losses including distributional similarity, orthogonal loss, reconstruction loss and task prediction loss to learn modality-invariant and modality-specific representation.

\noindent\textbf{MLF-DNN, MTFN, MLMF.} These are multi-task frameworks of the LF-DNN, TFN and LMF. All of them use independent unimodal annotations \cite{Yuetal2020}.

\noindent\textbf{Self-MM.} The Self-MM \cite{Yuetal2021} designs a unimodal label generation strategy based on the self-supervised method, then introduces unimodal subtasks to aid in learning modality-specific representations.

\subsection{Setup}

We run the models five times on each datasets and report the average performance on the test set. In the training procedure, we consider tuning the following hyper-parameters: learning rate (lr), batch size (bs), dropout (tdrp, adrp, vdrp) and  number of hidden units of each modality-specific subnetwork (thid, ahid, vhid), out dimensions of text subnetwork (tout), dropout and units of fusion layer (fdrp, fdim), attention heads and weight decay (wgd). The values for each hyperparameter of different datasets are shown in Table 2. To do a fair comparison on IEMOCAP, we also give four parameter settings corresponding to four emotions.

We use Adam optimizer with initial learning rate throughout all experiments and perform early stopping by 20 epochs. Besides, in Section 3, the values of $d_t$, $d_a$, $d_v$ are 768, 33, 709; $d_{t1}$, $d_{a1}$, and $d_{v1}$ refer to tout, ahid and vhid; $d$ and $d_m$ correspond to fdim; $\lambda$ is 3; the padding character is 0. L1 loss is used for CH-SIMS dataset, CMU-MOSI dataset and MOSEI dataset, while cross entropy loss for IEMOCAP dataset. The same as that literature \cite{Yuetal2020}, we record the experimental results on CH-SIMS in two forms: multi-class classification and regression. For multi-class classification, we report 2-class accuracy (Acc-2), 3-class accuracy (Acc-3), 5-class accuracy (Acc-5) and Weighted F1 score (F1). For regression, we report Mean Absolute Error (MAE) and Pearson correlation (Corr). For MOSI and MOSEI dataset, the metrics are 7-class accuracy (Acc-7), Acc-5, 2-class accuracy, Weighted F1 score, MAE and Corr. Different from the Acc-2 and F1 on CH-SIMS and IEMOCAP, following \cite{Yuetal2021}, the Acc-2 and F1-score are calculated in two ways: negative/non-negative (non-exclude zero) and negative/positive (exclude zero). For IEMOCAP dataset, we use Acc-2 and F1. For all metrics mentioned, except for MAE, the higher value means the better.

\begin{table}[width=.93\linewidth,cols=4,pos=h]
	\caption{Hyperparameters of BIMHA for the various tasks. For MOSI and MOSEI, the number on the left of / means the parameter for the aligned dataset, while the right one for the unaligned dataset.}\label{tbl2}
	\begin{tabular*}{\tblwidth}{@{}p{7mm}|p{5mm}p{7mm}p{9mm}|p{4mm}p{5mm}p{5mm}p{6mm}@{} }
		\toprule
	    \multirow{3}*{Para} & \multicolumn{7}{c}{Dataset}\\ \cline{2-8} 
		\multicolumn{1}{c|}{ } & \multirow{2}*{SIMS} & \multirow{2}*{MOSI} & \multirow{2}*{MOSEI} &  \multicolumn{4}{c}{IEMOCAP}\\ \cline{5-8} 
		\multicolumn{1}{c|}{ } & \multicolumn{3}{c|}{ } & H & S & A & N\\
		\midrule
		lr & 0.002 & 0.002 & 0.001 & 0.001 & 0.002 & 0.002 & 0.0003\\
		bs & 128 & 64/128 & 64/128 & 128 & 64 & 64 & 32\\
		tdrp & 0 & 0/0.2 & 0/0.1 & 0 & 0.5 & 0.5 & 0.15\\
		adrp & 0 & 0/0.2 & 0.1/0.1 & 0.3 & 0.15 & 0.2 & 0.2\\
		vdrp & 0 & 0/0.2 & 0.2/0.1 & 0.1 & 0.5 & 0.2 & 0\\
		thid & 128 & 64  & 64 & 64 & 256 & 64 & 128\\
		ahid & 16 & 64/16  & 64/32 & 8 & 32 & 8 & 16\\
		vhid & 128 & 64/8  & 64/32 & 16 & 4 & 8 & 4\\
		tout & 64 & 128 & 64 & 128 & 32 & 32 & 64\\
		fdrp & 0.2 & 0.2/0.1 & 0.2 & 0.15 & 0.5 & 0.1 & 0.2 \\
		fdim & 128 & 64/32 & 32/64 & 64 & 128 & 128 & 128 \\
	    heads & 6 & 4 & 8 & 4 & 6 & 6 & 8 \\
	    wgd & 0 & 0 & 0/0.001 & 0 & 0.001 & 0.001 & 0.001 \\
		\bottomrule
	\end{tabular*}
\end{table}

\section{Results and Analysis}
\subsection{Comparative Analysis}
Table 3 shows the comparative results on CH-SIMS dataset. The results of previous methods are published by the authors of \cite{Yuetal2020, Yuetal2021} with the link \footnote{https://github.com/thuiar/MMSA/blob/master/results/result-stat.md}. We can notice that our BIMHA outperforms other models in most metrics and shows significant improvement. The EF-LSTM has the worst performance. The reason may be that it does not fully learn the interaction information between modalities. Late fusion based method achieves certain improvement over EF-LSTM as it considers the intra-modal interaction first and extracts the more relevant information. A surprising discovery is that different from the previous studies where the order is MISA, MulT, DFG, MFN, LMF, TFN according to the performance from high to low, however, the order of that on CH-SIMS is opposite. TFN which introduces tensor fusion by calculating outer product from different modal features performs better than some models. MulT is not as good as expected than that of others, even though it employs the Transformer architecture. The performance of MISA is not satisfying. We suppose that this phenomenon may be subjected to the type of datasets. Recently, multi-task learning is widely applied in multimodal sentiment analysis to improve the generalization performance of multiple related tasks by utilizing the knowledge contained in different tasks. MLF-DNN, MTFN, MLMF and Self-MM are all multi-task learning based frameworks. So as expected they have indeed achieved better performance than that of the first eight methods. However, the performance of Self-MM is actually inferior to that of MLF-DNN, MTFN, MLMF. This is because the three methods are based on the unimodal labels manually annotated, while Self-MM is based on the unimodal labels generated by the continuous update of the model. And these generated labels may cause inaccuracy in predicting results. As our model considers the differences of interactions between pair-wise modalities and introduces bimodal attention via BMHA, the performance is largely superior to that of others especially on Acc-5. Besides, we provide the amount of parameters of each model to quantify the model complexity. Note M means million, K means thousand. As shown in Table 3, MTFN has a very large number of parameters due to its high dimensionality. The number of parameters of our model, 2.5 M, is less than that of TFN, DFG, MISA, MTFN and Self-MM.

In a word, the results suggest that combining intra-modal and inter-modal information can yield better performance. It also indicates that if fully mining features with high contribution and complementary information, the inter-bimodal interaction may benefit information fusion.
\begin{table}
	\caption{The results on CH-SIMS dataset for multimodal sentiment analysis, which have experienced paired t-test with p<0.05.}\label{tbl3}
	\begin{tabular*}{\tblwidth}{@{} LLLLCLLL@{} }
		\toprule
		Model & Acc-2 & Acc-3 & Acc-5 & F1 & MAE & Corr & Paras\\
		\midrule
		EF-LSTM\cite{Williamsetal2018} & 69.37 & 54.27 & 21.23 & 56.82 & 0.59 & 0.055 & 215K \\
		LF-DNN\cite{Yuetal2020} & 77.02 & 64.33 & 39.74 & 77.27 & 0.446 & 0.555 & 635K\\
		TFN\cite{Zadehetal2017} & 78.38 & 65.12 & 39.3 & 78.62 & 0.432 & 0.591 &35M \\
		LMF\cite{Liuetal2018} & 77.77 & 64.68 & 40.53 & 77.88 & 0.4412 & 0.576 & 1M\\
		MFN\cite{Zadehetal2018_a} & 77.9 & 65.73 & 39.47 & 77.88 & 0.435 & 0.582 &601K\\
		DFG\cite{Zadehetal2018_b} & 78.77 & 65.65 & 39.82 & 78.21 & 0.445 & 0.578 &2.7M\\
		MulT\cite{Tsaietal2019_b} & 78.56 & 64.77 & 37.94 & 79.66 & 0.453 & 0.564 &1.8M \\
		MISA\cite{Hazarikaetal2020}& 69.45 & 54.14 & 21.79 & 57.02 & 0.588& 0.142 &123M \\
		MLF-DNN\cite{Yuetal2020} & 80.44 & \textbf{69.37} & 40.22 & 80.28 & 0.396 & 0.665 &283K\\
		MTFN\cite{Yuetal2020} & 81.09 & 68.80 & 40.31 & 81.01 & 0.395 & \textbf{0.666} & 140M \\
		MLMF\cite{Yuetal2020} & 79.34 & 68.36 & 41.05 & 79.07 & 0.409 & 0.639 &1.4M\\
		Self-MM\cite{Yuetal2021} & 80.04 & 65.47 & 41.53 & 80.44 & 0.425 & 0.595 &102M\\
		BIMHA & \textbf{82.71} & 69.23 &\textbf{45.21} &\textbf{82.72} &\textbf{0.385} &0.66 &2.5M\\
		\hline
		T-test & 0.001 &0.0018 &0.0071 & 0.0008 & 0.0115 & 0.038&   \\
		\bottomrule
	\end{tabular*}
\end{table}

\subsection{Ablation Study}
To examine the functionality of the overall architecture and the effect of different combinations of bimodal features, we conduct experiments from two aspects. One is the combinations of single modality (Table 4), the other is the combinations of pair-wise modalities (Table 5). For each one, the experiments are divided into three groups. The first group represents the results of single modality / pair-wise modalities, the second and third groups are the results of every two or more combinations. Moreover, except for the third group using the BMHA, others adopt simple concatenation followed by Fully Connected layers (FC). The experimental results are shown in Table 4 and Table 5. Bold numbers denote the best results in each group. UM means the unimodal features are used in BMHA, while BM means the bimodal features (outer product between modalities) are used in BMHA. Besides, two elements correspond to two-MHA based BM while three elements correspond to three-MHA based BM. Experiment 11 in Table 4 also represents BIMHA without IMI, Experiment 7 in Table 5 represents BIMHA without IBI.

In Table 4, from the results of experiments 1, 2, 3, we can infer that text is the most predictive, while acoustic modality performs poorly. This may deviate from our common sense where the phoneme, pitch and other features are often expressive. We argue that most of the features extracted from the raw audio and video data are handcrafted and the temporal information is not fully considered, while the rich semantic information in text is obtained by using the advanced pre-trained model. For the combinations of any two modalities, the results of 4, 5, 6, 8, 9, 10 show that the visual modality combined with text can provide more information than other combinations. However, three modalities achieve the best performance. When the single modality is endowed with the attention via BMHA, the combination of acoustic and visual modality and the combination of visual and textual modality obtain improvement by a large margin. However, it is not as good as expected in terms of A+T and A+V+T. It is because the modalities are not aligned at the word level and the audio contains noises like background music.

In Table 5, the experiments are based on the bimodal features calculated via outer product. As we can see, VT performs relatively well. It also enhances the performance when combining with AT and AV respectively. Compared the results of 4, 5, 6 with those of 1, 2, 3, it is easy to figure out that the combinations of two pair-wise modalities can generally improve the performance. AT + VT achieves the best results. It is worth noting that the combination of three pair-wise modalities (7) is not as excellent as that of the two, because the simple concatenation fails to consider the different contributions of them. With BMHA, the model's performance improves a lot. The results of 8, 9, 10, 11 show that BMHA with three-MHA, namely AV+AT+VT (BM) is more predictable than that with two-MHA.

To further analyze the effect of using unimodal features and bimodal features, we compare the results in Table 4 with those in Table 5 group by group. In the first group, bimodal fusion is more efficient to some extent. However, different combinations take different influences. As we can see, text fused with visual modality achieves the significant performance. However, with acoustic modality, the performance of AT is even worse than that of single textual modality. In the second group, the combinations of any two pair-wise modalities are more efficient than the concatenation of any two single modalities. Since the interactions among AV, AT, VT are different, the direct concatenation of them is unable to obtain such differences. As a result, the performance of AV+AT+VT is not better than that of A+V+T. Take these differences into consideration, BMHA conducts interactions between any two pair-wise modalities, namely AT-AV, AT-VT, AV-VT, and allocates bimodal attention. The results in the third group demonstrate that BMHA makes promising improvements. Bimodal fusion enhances the function of BMHA and bimodal attention indeed works. That's why we perform bimodal fusion first rather than trimodal fusion. Furthermore, from the results of 11 (no IMI) in Table 4 and 7 (no IBI), 11 (full) in Table 5, it is obvious that the module of IBI is more important than IMI. The full model with IBI and IMI achieves the best performance.

Specially, we have studied the effect of combining bimodal features with unimodal features. The results are shown in Table 6. It is obvious that VT combined with A performs better than others do. However, compared the results of 4, 5, 6 with those of 1, 2, 3, we find that the performance of BMHA taking both bimodal and unimodal features as the target information is not satisfying. The reason may be due to the differences between bimodal representation and unimodal representation. The model can hardly extract the consistent information. As a whole, it is better to use combinations rather than a single modality or single pair-wise modalities. Besides, bimodal information indeed enhances the model's performance.

\begin{table}[width=.9\linewidth,cols=4,pos=h]
	\caption{The experimental results on different combinations of single modalities. UM means it uses BMHA with single modality as target information, others use concatenation followed by FC. Bold numbers denote the best results in each group.}\label{tbl4}
	\begin{tabular*}{\tblwidth}{@{} LLCCCCCC@{} }
		\toprule
	Num&	Model & Acc-2 & Acc-3 & Acc-5 & F1 & MAE & Corr\\
		\midrule
	1&	A & 69.37 & 54.27 & 21.23 & \textbf{81.91} & 0.590 & 0.032\\	
	2&	V & 75.67 & 57.94 & 26.7 & 76.82 & 0.517 & 0.455\\	
	3&	T & \textbf{76.5} & \textbf{56.24} & \textbf{35.1} & 77.44 & \textbf{0.471} & \textbf{0.508}\\
		\hline
	4&	A+V & 74.14 & 56.85 & 33.35 & 76.18 & 0.504 & 0.425\\	
	5&	A+T & 75.89 & 59.52 & 34.62 & 79.58 & 0.485 & 0.446\\	
	6&	V+T & 77.64 & 62.06 & 34.18 & 78.14 & 0.461 & 0.550\\	
	7&	A+V+T & \textbf{79.91} & \textbf{66.78} & \textbf{42.14} & \textbf{80.09} & \textbf{0.41} & \textbf{0.616}\\
		\hline	
		\hline
	8&	A+V(UM) & 76.63 & 56.19 & 34.44 & 78.01 & 0.481 & 0.472\\	
	9&	A+T(UM) & 75.36 & 60.04 & 33.9 & 78.74 & 0.493 & 0.47\\	
	10&	V+T(UM) & 79.26 & 63.46 & 40.88 & 79.76 & 0.438 & 0.581\\	
	11&	A+V+T(UM) & 	\textbf{79.52} & \textbf{66.74} & \textbf{41.27} & \textbf{79.95} & \textbf{0.421} & \textbf{0.611}\\
		\bottomrule
	\end{tabular*}
\end{table}

\begin{table}[width=.9\linewidth,cols=4,pos=h]
	\caption{The experimental results on different combinations of pair-wise modalities. BM means it uses BMHA with pair-wise modalities as target information, others use concatenation followed by FC. BIMHA corresponds AV+AT+VT (BM).}\label{tbl5}
	\begin{tabular*}{\tblwidth}{@{} LLCCCCCC@{} }
		\toprule
		Num & Model & Acc-2 & Acc-3 & Acc-5 & F1 & MAE & Corr\\
		\midrule
	1&	AV & 74.31 & 58.6 & 32.3 & 75.68 & 0.498 & 0.446\\	
	2&	AT & 70.42 & 54.22 & 21.31 & \textbf{81.31} & 0.583 & 0.119\\	
	3&	VT & \textbf{78.91} & \textbf{61.84} & \textbf{39.87} & 79.76 & \textbf{0.438} & \textbf{0.581}\\
		\hline
	4&	AT+VT & \textbf{79.56} & \textbf{64.07} & \textbf{41.71} & \textbf{80.5} & \textbf{0.43} & \textbf{0.572}\\	
	5&	AT+AV & 77.94 & 60.26 & 36.19 & 78.58 & 0.452 & 0.536\\	
	6&	VT+AV & 79.3 & 63.37 & 36.81 & 80.08 & 0.444 & 0.568\\	
	7&	AV+AT+VT & 	77.59 & 63.11 & 37.64 & 78.49 & 0.444 & 0.555\\
		\hline
		\hline
	8&	AT+VT(BM) & 79.65 & 64.03 & 38.82 & 80.76 & 0.431 & 0.595\\	
	9&	AT+AV(BM) & 78.56 & 61.49 & 37.55 & 79.51 & 0.449 & 0.58\\	
	10&	VT+AV(BM) & 79.74 & 65.95 & 42.71 & 80.59 & 0.421 & 0.607\\	
	11&	BIMHA& \textbf{82.71} & \textbf{69.23} &\textbf{45.21} &\textbf{82.72} &\textbf{0.385} &\textbf{0.66}\\
		\bottomrule
	\end{tabular*}
\end{table}

\begin{table}[width=.9\linewidth,cols=4,pos=h]
	\caption{The experimental results on different pair-wise modalities with the single modality. HM means it uses BMHA with hybrid modalities in bimodal and unimodal.}
	\label{tbl6}
	\begin{tabular*}{\tblwidth}{@{} LLCCCCCC@{} }
		\toprule
		Num & Model & Acc-2 & Acc-3 & Acc-5 & F1 & MAE & Corr\\
		\midrule
		1&	AV+T & 78.91 & 64.73 & 39.39 & 79.43 & 0.435 & 0.573\\	
		2&	AT+V & 79.39 & 63.89 & 37.33& 80.02 & 0.434 & 0.585\\	
		3&	VT+A & \textbf{79.96} & \textbf{65.3} & \textbf{40.83} & \textbf{80.59} & \textbf{0.412} & \textbf{0.624}\\
		\hline
		4&	AV+T(HM) & 77.77 & 60.57 & 37.99 & 78.84 & 0.443 & 0.559\\	
		5&	AT+V(HM) & 77.99 & 60.09 & 36.72 & 78.84 & 0.446 & 0.568\\	
		6&	VT+A(HM) & \textbf{79.87} & \textbf{64.73} & \textbf{42.71} & \textbf{81.1} & \textbf{0.418} & \textbf{0.607}\\	
		\bottomrule
	\end{tabular*}
\end{table}

\subsection{The Generalization of Model}
To verify the generalization of the proposed model, we conduct experiments on the CMU-MOSI, MOSEI and IEMOCAP English datasets. For the CMU-MOSI and MOSEI, we evaluate the model on the word-aligned datasets (BIMHA$_1$) and the word-unaligned datasets (BIMHA$_2$). The MCTN \cite{Phametal2019}, RAVEN \cite{Wangetal2019}, CIA \cite{Chauhanetal}, MMUU-BA \cite{Ghosaletal2018} are also referred as baselines. The first two are efficient methods for the aligned dataset. For the unaligned dataset, they need to use connectionist temporal classification (CTC) \cite{Gravesetal2006} as auxiliary. Others are kind of hierarchical fusion methods, where CIA learns the contributing utterances in the neighborhood by exploiting the interaction among the input modalities, MMUU-BA is an utterance-level method which also introduces bimodal attention. The results are shown in Table 7. The bold number indicates the best results on the datasets. Compared with aligned models except for CIA, our model achieves significant improvements in most evaluation metrics, especially on MOSEI dataset. CIA is more powerful on MOSI dataset. However, compared with unaligned models, our model outperforms LF-DNN, TFN, LMF, MCTN, RAVEN, but it is not as competitive as MulT, MISA and Self-MM in some metrics. The gap is even more obvious on the MOSI dataset due to the small size of this dataset. It is expected that Self-MM gets better results as it uses fine-grained labels and multi-task learning, which can help better learn the informative features in single modality on unaligned data. Generally, models using aligned corpus can get better results. However, in our experiments, we find that the performance of BIMHA is better on unaligned dataset through the comparison between BIMHA$_1$ and BIMHA$_2$.  Since the main motivation of our model is for Chinese multi-modal sentiment analysis, and the CH-SIMS dataset only contains unaligned data, the model is more sensitive to unaligned data.
\begin{table*}[width=2.5\linewidth,cols=4,pos=h]
	\centering
	\caption{The experimental results on MOSI and MOSEI (bold number means the best results on the dataset. For Acc-2 and F1, the number \newline on the left of / denotes "negative/non-negative" and  the right is "negative/positive").}\label{tbl7}
	\scalebox{0.8}{
	\begin{tabular*}{\tblwidth}{@{} lllllll|lllllll@{} }
	\cline{1-14}
	\multicolumn{1}{l}{\multirow{2}{*}{Model}} & \multicolumn{6}{c|}{MOSI}                                  & \multicolumn{6}{c}{MOSEI}                                 & \multirow{2}{*}{Data Setting} \\ \cline{2-13}
	\multicolumn{1}{c}{}                       & Acc-7 & Acc-5 & \makecell[c]{Acc-2}       & \makecell[c]{F1}          & MAE   & Corr  & Acc-7 & Acc-5 & \makecell[c]{Acc-2}       & \makecell[c]{F1}          & MAE   & Corr  &                               \\ [3pt] 	\cline{1-14}
	EF-LSTM\cite{Williamsetal2018}                                    & 35.9  & 40.15 & 77.38/78.48 & 77.35/78.51 & 0.949 & 0.669  & 50.01 & 51.16 & 77.84/80.79 & 78.34/80.67 & 0.601 & 0.683 & Aligned                       \\
	MFN\cite{Zadehetal2018_a}                                       & 35.83 & 40.47 & 77.67/78.87 & 77.63/78.90 &  0.927&  0.670 & 51.34 & 52.76 & 78.94/82.86 & 79.55/82.85 & 0.573 & 0.718 & Aligned                       
	\\
	DFG\cite{Zadehetal2018_b}                                  & 34.64 & 38.63 & 77.14/78.35 & 77.08/78.35 & 0.956 & 0.649 & 51.37 & 52.69 & 81.28/83.48 & 81.48/83.23 & 0.575 & 0.713 & Aligned                       \\
	MCTN\cite{Phametal2019}   & 35.6 & - & 79.3 & 79.1 & \textbf{0.909} & 0.677 & 49.6 & - & 79.8 & 80.6 & 0.609 & 0.67 & Aligned
	\\
	RAVEN\cite{Wangetal2019} & 33.2 & - & 78 & 76.6 & 0.915 &\textbf{0.691} & 50.0 &- & 79.1 & 79.5 &0.614 & 0.662 & Aligned
	\\
	MMUU-BA\cite{Ghosaletal2018} & 33.8 & - & 78.2 & 78.1 & 0.947 &0.675 & 48.4 &- & 80.7 & 80.2 &0.627 & 0.672 & Aligned
	 \\ 
	CIA\cite{Chauhanetal} & \textbf{38.92}&-&79.88&79.54&0.914&0.689 & 50.14&49.15&80.37&78.23&0.683&0.594&Aligned\\
	MulT\cite{Ghosaletal2018} & 33.6 & - & 78.7 & 78.4 & 0.964 &0.662 & 46.6 &- & 80.2 & 79.8 &0.657 & 0.661 & Aligned
	\\        
	BIMHA$_1$                                   & 35.86 & \textbf{40.82} & 78.57/\textbf{80.18} &  78.55/\textbf{80.23} & 0.9291 & 0.6633 &  \textbf{52.69} &  \textbf{54.17} &  \textbf{83.19/83.93} &  \textbf{83.21/83.64} &  \textbf{0.562} &  \textbf{0.729} & Aligned                       
	\\
	\cline{1-14}
	LF-DNN\cite{Yuetal2020}                                     & 34.52 & 38.05 & 77.52/78.63 & 77.46/78.63 & 0.955 & 0.658 & 50.83 & 51.97 & 80.6/82.74  & 80.85/82.52 & 0.58 & 0.709 & Unaligned                     \\
	TFN\cite{Zadehetal2017}                                       & 34.46 & 39.39 & 77.99/79.08 & 77.95/79.11 & 0.947 & 0.673 & 51.6  & 53.1  & 78.50/81.89 & 78.96/81.74 & 0.573 & 0.714 & Unaligned                     \\
	LMF\cite{Liuetal2018}                                        & 33.82 & 38.13 & 77.9/79.18  & 77.8/79.15  & 0.950 & 0.651  & 51.59 & 52.99 & 80.54/83.48 & 80.94/83.36 & 0.576 & 0.717 & Unaligned                     \\
	CTC+MCTN   & 32.7 & - & 75.9 & 76.4 & 0.991 & 0.613 & 48.2 & - & 79.3 & 79.7 & 0.631 & 0.645 &Unaligned
\\
CTC+RAVEN & 31.7 & - & 72.7 & 73.1 & 1.076 &0.544 & 45.5 &- & 75.4 & 75.7 &0.664 & 0.599 & Unaligned
\\           
	MulT\cite{Tsaietal2019_b}                                       & 36.91 & 42.68 & 79.71/80.98 & 79.63/80.95 & 0.880 & 0.702 & 52.84 & 54.18 & 81.15/84.63 & 81.56/84.52 & 0.559 & 0.733 & Unaligned                     \\
	MISA\cite{Hazarikaetal2020}                                       & 41.37 & 47.08 & 81.84/83.54 & 81.82/83.58 & 0.777 & 0.778 & 52.05 & 53.63 & 80.67/84.67 & 81.12/84.66 & 0.558 &0.752 & Unaligned                     \\
	Self-MM\cite{Yuetal2021}                                    & \textbf{46.67} &\textbf{53.47} & \textbf{83.44}/\textbf{85.46} & \textbf{83.36}/\textbf{85.43} & \textbf{0.708} & \textbf{0.796} & \textbf{53.87} & \textbf{55.53} & 83.76/\textbf{85.15} &\textbf{83.82}/\textbf{84.90} & \textbf{0.531} & \textbf{0.765} & Unaligned                     \\
	BIMHA$_2$                                   & 36.44 & 43.29 & 78.57/80.3  & 78.5/80.03  & 0.925 & 0.671 & 52.11 & 53.36 & \textbf{84.07}/83.96 & 83.35/83.5  & 0.559 & 0.731 & Unaligned                     \\	\cline{1-14}
	\end{tabular*}}
\end{table*}

Table 8 reports the results on IEMOCAP dataset. The MFM \cite{Tsaietal2019} consists of a discriminative model for prediction and a generative model for reconstructing the input data. The ICCN \cite{Sunetal2020} uses canonical correlation to analyze hidden relationships between text, audio, and video. The results show that our model achieves the best performance on F1 of all emotions. The results on Acc-2 of Happy and Angry emotions are slightly lower than those of ICCN. Besides, we can see BIMHA's competitive performance in predicting Neutral emotion. Overall, all models achieve the best results on Happy emotion and the worst ones on Neutral emotion. This may be related to the unbalanced distribution of data in the dataset.
\begin{table*}[width=1\linewidth,cols=4,pos=h]
	\caption{The experimental results on IEMOCAP.}\label{tbl8}
	\centering
	\begin{tabular*}{\tblwidth}{@{} lllllllll@{} }
		
	\cline{1-9}
		& \multicolumn{8}{c}{IEMOCAP}                                                                                                                                                                               \\ \cline{1-9}
		\multirow{2}{*}{ } & \multicolumn{2}{c}{Happy}         & \multicolumn{2}{c}{Angry}                             & \multicolumn{2}{c}{Sad}                               & \multicolumn{2}{c}{Neutral}                           \\ \cline{2-9}
		& \multicolumn{1}{c}{Acc-2} & \makecell[c]{F1}    & \multicolumn{1}{c}{Acc-2} & \makecell[c]{F1}                        & \multicolumn{1}{c}{Acc-2} & \makecell[c]{F1}                        & \multicolumn{1}{c}{Acc-2} & \makecell[c]{F1}                        \\ 	\cline{1-9}
		TFN\cite{Zadehetal2017}                       & 86.66                     & 83.6  & 87.11                     & \multicolumn{1}{l}{87.03} & 85.64                     & \multicolumn{1}{l}{85.75} & 68.9                      & \multicolumn{1}{l}{68.03} \\
		LMF\cite{Liuetal2018}                       & 86.14                     & 85.8  & 86.24                     & \multicolumn{1}{l}{86.41} & 84.33                     & \multicolumn{1}{l}{84.40} & 69.62                     & \multicolumn{1}{l}{68.75} \\
		MFM\cite{Tsaietal2019}                       & 86.67                     & 84.66 & 86.99                     & \multicolumn{1}{l}{86.72} & 85.67                     & \multicolumn{1}{l}{85.66} & 70.26                     & \multicolumn{1}{l}{69.98} \\
		ICCN\cite{Sunetal2020}                      & \multicolumn{1}{l}{\textbf{87.41}} & 84.72 & \multicolumn{1}{l}{\textbf{88.62}} & 88.02                     & \multicolumn{1}{l}{86.26} & 85.93                     & \multicolumn{1}{l}{69.73} & 68.47                     \\
		BIMHA                     & \multicolumn{1}{l}{86.57} & \textbf{85.8}  & \multicolumn{1}{l}{88.27} & \textbf{88.41}                     & \multicolumn{1}{l}{\textbf{86.57}} & \textbf{86.26}                     & \multicolumn{1}{l}{\textbf{72.17}} & \textbf{71.25}                     \\ \cline{1-9}	
	\end{tabular*}
\end{table*}

\subsection{Data Visualization}
\begin{figure}
	\centering
	\includegraphics[scale=.4]{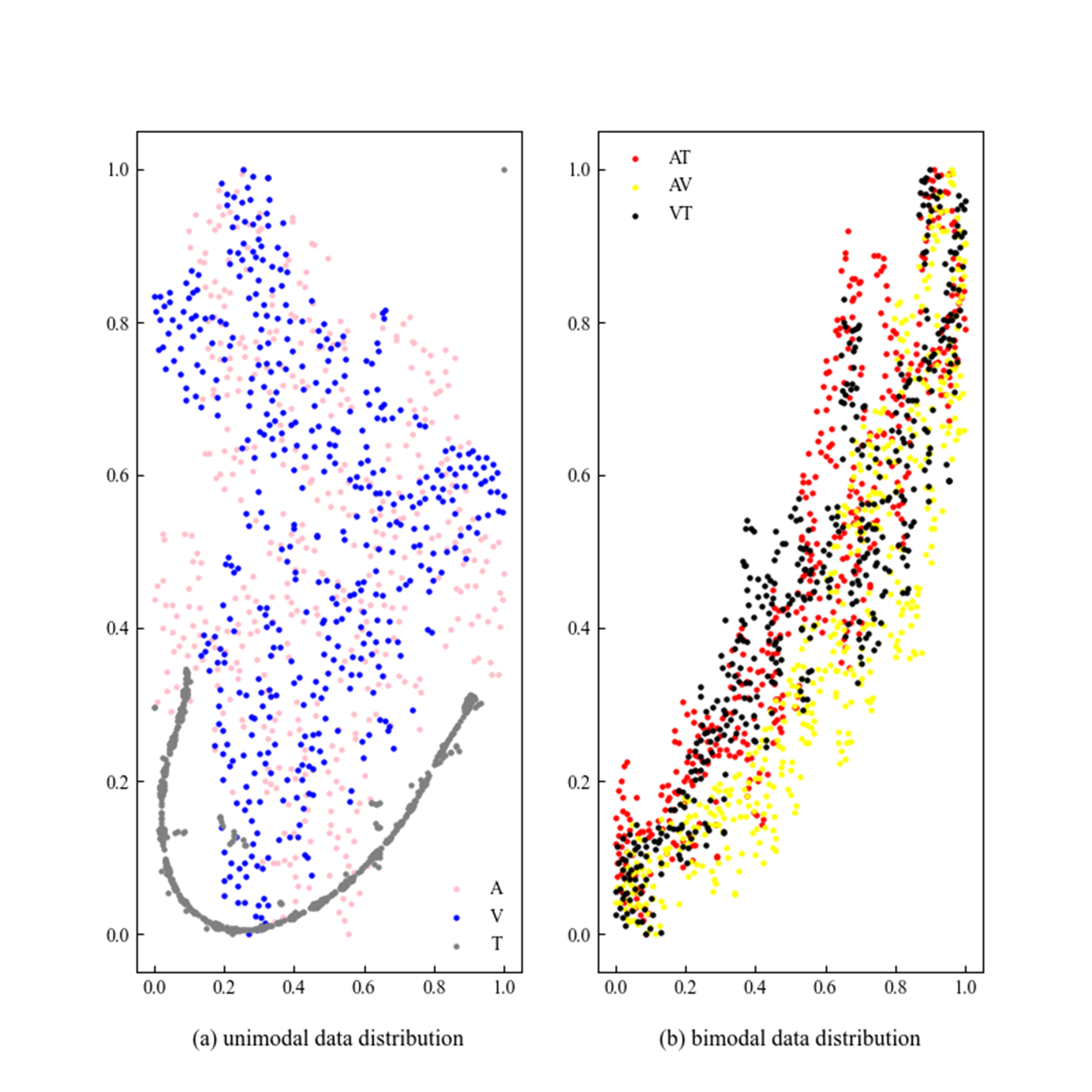}
	\caption{The distribution of unimodal features and bimodal features.}
	\label{FIG:3}
\end{figure}

\begin{figure}
	\centering    
	\subfigure[Bimodal attention on the samples of test dataset.] 
	{
		\begin{minipage}{0.5 \textwidth}
			\centering
			\includegraphics[width=6cm]{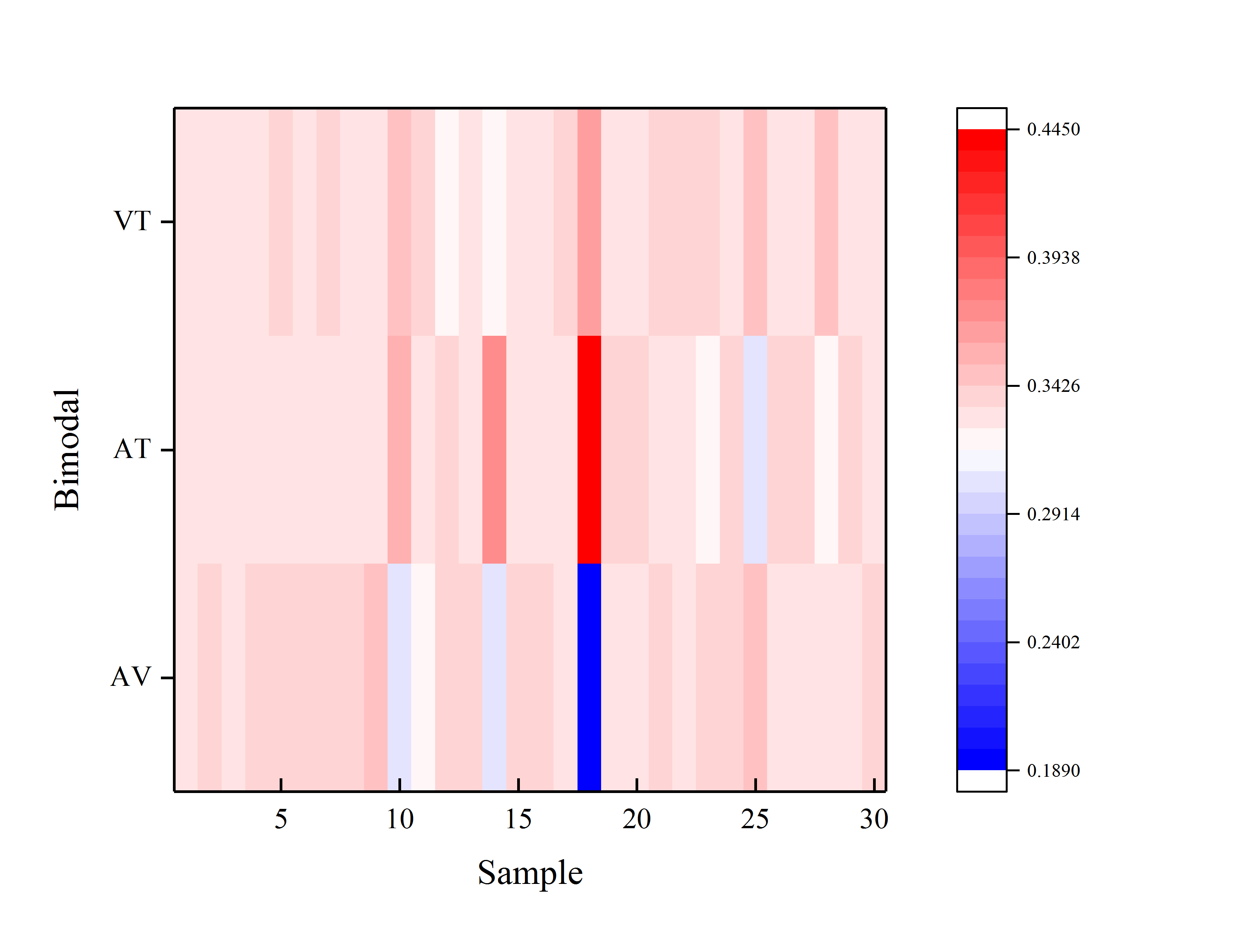}\vspace{4pt}
		\end{minipage}
	}
	\subfigure[Examples from the CH-SIMS dataset.]
	{
		\begin{minipage}{0.5 \textwidth}
			\centering     
			\includegraphics[width=6cm]{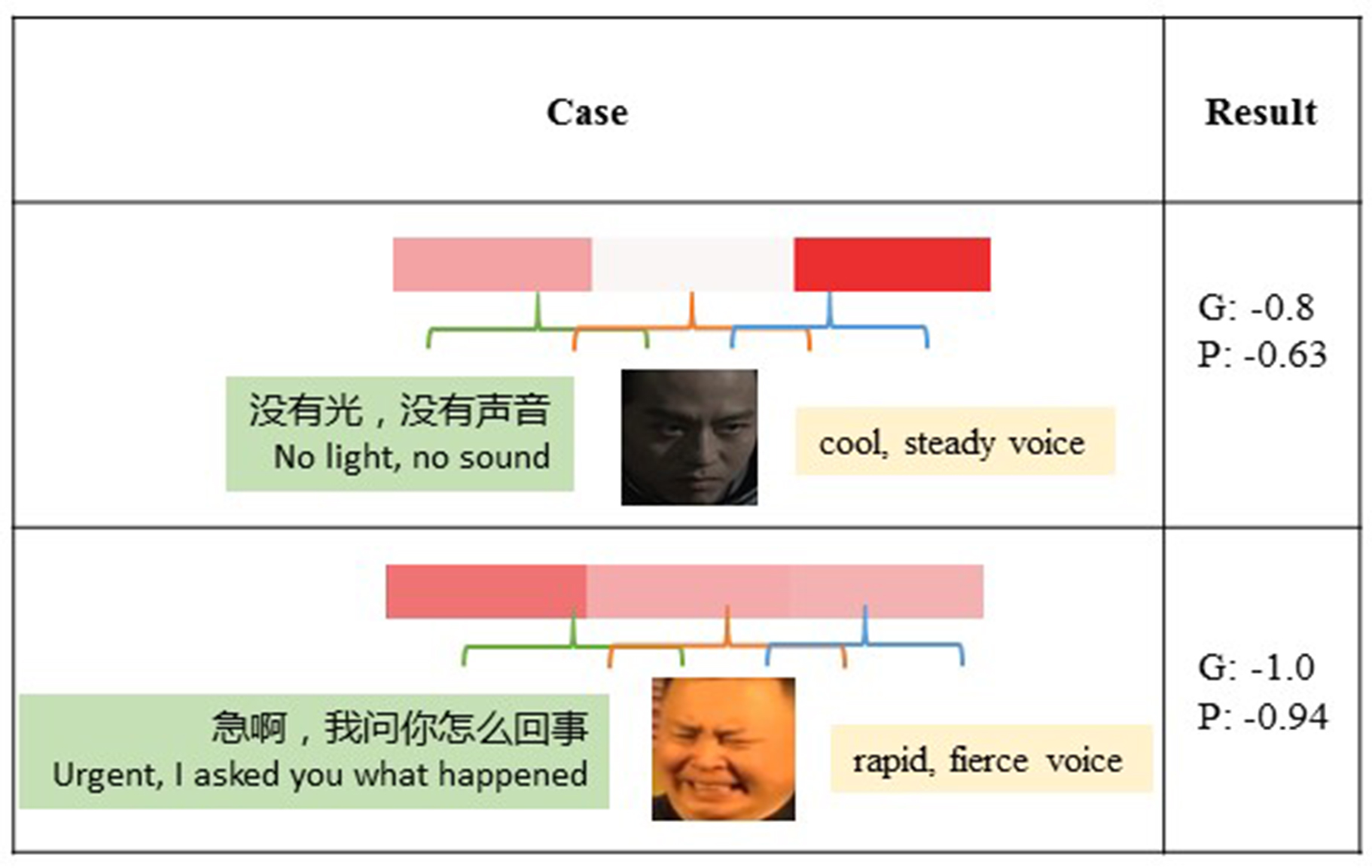}
		\end{minipage}
	}
	\caption{Attention visualization. Figure (a) demonstrates the bimodal attention on some samples of the test dataset, and Figure (b) shows the prediction results of the proposed model on two examples. G means ground truth and P is the prediction result.}
	\label{Fig:4}  
\end{figure}
In addition, to illustrate intuitively how the model works, we use T-SNE \cite{vanetal2008} to visualize the distribution of unimodal feature representations and bimodal feature representations with assigned attention in a two-dimensional coordinate system. The horizontal and vertical axes correspond to their position coordinates. In Figure 3, Figure 3(a) shows the unimodal feature distribution. It is obvious that except for the compact and regular representation of the textual features, the feature representations of audio and video are relatively scattered, and the interaction between them is relatively small. Figure 3(b) presents the bimodal feature distribution with bimodal attention. We can see that the distributions of all the bimodal features are relatively consistent. We think the reasons can be explained from two aspects. One is due to the network's approximation capability, and the second may owe to the weighted strategies which can better balance these features. That is to say, similar to the trimodal systems, it is effective as the different modalities help each other to provide complementary information for sentiment analysis. Our model may learn more informative features rather than noisy features.

We take 30 samples of the test set and visualize the corresponding bimodal attention on these samples. As shown in Figure 4(a), each column shows the attention scores corresponding to AV, AT and VT of a sample. The color ranging from blue to red means the increasing of score. We can see that the importance of AV, AT and VT on single sample and on the whole dataset are different. In this case, the overall trend is that the bimodal features combining text and video have more influence on the effectiveness of multimodal fusion. In other words, the contribution of certain pair-wise modalities is greater on the entire dataset, while for each sample, the contribution of that is variable. Therefore, when performing sentiment analysis, we should consider the degree of information contribution after the fusion of various modalities. Figure 4(b) presents two examples where our model predicts the sentiment properly by considering both inter-modal and inter-bimodal interaction. In case 1, we can see that the model provides higher scores on the features combining audio and video, but attaches less importance to AT features. In case 2, VT features are more important and can help the model better learn the man's sentiment state. The results show our method can better fuse the different modalities and explore the key information.

\section{Conclusion}
Considering that the difference of the interaction between multiple modalities makes the information contribution different when performing multimodal fusion, this paper proposes a sentiment analysis method based on bimodal information-augmented multi-head attention. Following this method, the independent and consistent information of bimodal features are obtained via BMHA, which can benefit the subsequent information fusion. We have evaluated our approach on CH-SIMS, CMU-MOSI, MOSEI and IEMOCAP datasets, and intuitively explained the principle of the model and the bimodal attention. The results show that the proposed model is effective and the performance is better than that of the existing models.

In future, we will use more relevant fusion methods to conduct experiments on the CH-SIMS dataset so as to provide more benchmark results for Chinese-based multimodal sentiment analysis. In addition, we will try to improve the model following the recent trends and explore what interesting results could be obtained by adding gender, emotional duration, and other dimensional characteristics.

\section*{Acknowledgments}
The authors would like to thank the funding from the Open Project Program of Shanghai Key Laboratory of Data Science (No. 2020090600004) and the resources and technical support from the High Performance Computing Center of Shanghai University, and Shanghai Engineering Research Center of Intelligent Computing System (No. 19DZ2252600).

\printcredits

\bibliographystyle{cas-model2-names}

\bibliography{bimha_clean}



\end{sloppypar}
\end{document}